\documentclass{trbunofficial}
\usepackage{graphicx}

\usepackage[hidelinks]{hyperref}
\usepackage{array}
\usepackage{algorithm}
\usepackage{algorithmic}
\usepackage{amsmath}
\usepackage{url}

\begin{document}

\begin{flushleft}
    \Large \textbf{Real-World Data Inspired Interactive Connected Traffic Scenario Generation} \\[24pt]
    
    \normalsize
      \textbf{Junwei You}\\
      University of Wisconsin-Madison\\
      Email: jyou38@wisc.edu\\
      \hfill\break
      \textbf{Pei Li, Ph.D., Corresponding Author}\\
      University of Wisconsin-Madison\\
      Email: pei.li@wisc.edu\\
      \hfill\break%
      \textbf{Yang Cheng, Ph.D.}\\
      University of Wisconsin-Madison\\
      Email: cheng8@wisc.edu\\
      \hfill\break%
      \textbf{Keshu Wu, Ph.D.}\\
      University of Wisconsin-Madison\\
      Email: kwu84@wisc.edu\\
      \hfill\break%
      \textbf{Rui Gan}\\
      University of Wisconsin-Madison\\
      Email: rgan6@wic.edu\\
      \hfill\break%
      \textbf{Steven T. Parker, Ph.D.}\\
      University of Wisconsin-Madison\\
      Email: sparker@engr.wisc.edu\\
      \hfill\break
      \textbf{Bin Ran, Ph.D.}\\
      University of Wisconsin-Madison\\
      Email: bran@wisc.edu
\end{flushleft}

\newpage
\section{Abstract}
Simulation is a crucial step in ensuring accurate, efficient, and realistic Connected and Autonomous Vehicles (CAVs) testing and validation. As the adoption of CAV accelerates, the integration of real-world data into simulation environments becomes increasingly critical. Among various technologies utilized by CAVs, Vehicle-to-Everything (V2X) communication plays a crucial role in ensuring a seamless transmission of information between CAVs, infrastructure, and other road users. However, most existing studies have focused on developing and testing communication protocols, resource allocation strategies, and data dissemination techniques in V2X. There is a gap where real-world V2X data is integrated into simulations to generate diverse and high-fidelity traffic scenarios. To fulfill this research gap, we leverage real-world Signal Phase and Timing (SPaT) data from Roadside Units (RSUs) to enhance the fidelity of CAV simulations. Moreover, we developed an algorithm that enables Autonomous Vehicles (AVs) to respond dynamically to real-time traffic signal data, simulating realistic V2X  communication scenarios. Such high-fidelity simulation environments can generate multimodal data, including trajectory, semantic camera, depth camera, and bird's eye view data for various traffic scenarios. The generated scenarios and data provide invaluable insights into AVs' interactions with traffic infrastructure and other road users. This work aims to bridge the gap between theoretical research and practical deployment of CAVs, facilitating the development of smarter and safer transportation systems.

\hfill\break%
\noindent\textit{Keywords}: Vehicle-to-Everything, Connected and Autonomous Vehicles, Singal Phase and Timing, High-Fidelity Traffic Simulation
\newpage

\section{Introduction}
The development of Connected and Autonomous Vehicles (CAVs) is revolutionizing the transportation industry by integrating advanced perception, control, and communication technologies. In particular, Vehicle-to-Everything (V2X) communication enables CAVs to interact with other road users and with surrounding infrastructure, such as traffic signals and road signs \cite{huang2024toward}. V2X aims to enhance road safety, optimize traffic flow, and reduce environmental impact \cite{hobert2015enhancements, you2024v2x}. Among the different forms of V2X, Vehicle-to-Infrastructure (V2I) communication, facilitated by Roadside Units (RSUs), plays a crucial role particularly.

\subsection{V2X Simulation}
The development of V2X communication technologies and their integration with AVs has led to a variety of simulation frameworks aimed at testing and refining these systems. These frameworks are crucial for understanding how V2X and AV technologies can be effectively deployed in real-world environments. For example, one of the foundational studies in this area \cite{fu2021network} presents a simulation platform that integrates network and driving scenarios to assess the impact of Cellular Vehicle-to-Everything (C-V2X) technology on vehicle communication and control. This platform integrates network and driving scenarios using SUMO and CARLA simulators, which enables the evaluation of road traffic and vehicle dynamics under a C-V2X framework. Another study \cite{choudhury2016integrated} focuses on simulating vehicle platooning, demonstrating the potential of V2X to enhance traffic efficiency by coordinating vehicle movements. The importance of integrating V2X sensors into existing simulation tools is highlighted in \cite{grimm2024co}. This study expands the CARLA simulator's capabilities by enabling detailed simulations of V2X interactions, particularly in urban environments where complex traffic scenarios are common. Meanwhile, \citet{lee2019building} underscores the necessity for simulation frameworks that can accommodate various V2X communication protocols, ensuring comprehensive testing of system interoperability. 

Performance analysis is a critical component of these studies, as seen in \cite{eckermann2019performance} which evaluates the communication protocol's reliability and efficiency. Similarly,~\citet{chen2021performance} provides a detailed examination of latency and throughput, essential metrics for assessing the feasibility of deploying V2X systems in densely populated urban areas. The findings highlight the critical role of media access control (MAC) layer parameters, such as resource reselection probability and channel bandwidth, in influencing packet reception ratios and overall communication reliability. To conduct performance analysis, specialized simulation tools play a vital role. For example, \citet{zhang2020v2xsim} offers a platform specifically designed to test V2X applications like traffic signal prioritization, providing valuable insights into how these systems can be optimized for real-world use. The emphasis on realistic data in simulations is also highlighted by \cite{queck2008realistic} which addresses the challenges of simulating V2X communications under variable environmental conditions. In addition, comprehensive simulation environments offer a holistic approach that is crucial for evaluating V2X protocols and their effectiveness in improving traffic flow and safety~\cite{schunemann2011v2x}. They integrate multiple aspects of traffic and network modeling, facilitating the development of advanced traffic management systems. Lastly, \citet{qom2016evaluation} explores how V2X technologies can enhance the capabilities of AVs, particularly through applications like cooperative adaptive cruise control. This study highlights the synergy between V2X and AV technologies, demonstrating how they can be integrated to create more efficient and safer transportation systems.

These papers collectively provide a comprehensive overview of the current state of V2X simulation and its critical role in the development of CAV technologies. They underscore the importance of versatile and high-fidelity simulation tools in evaluating these systems' performance and scalability, paving the way for more intelligent and adaptive traffic management solutions.

\subsection{RSU Data Utilization}
The integration of Roadside Units (RSUs) in vehicular networks is pivotal for enhancing V2X communication and data services, which are critical for the development of CAVs. The literature reveals various strategies and frameworks focused on optimizing RSU data utilization, addressing key aspects such as data scheduling, dissemination, and resource allocation. Specifically, efficient data scheduling is a recurrent theme across several studies, which highlights the importance of managing the flow of information between vehicles and RSUs. \citet{mershad2012score} proposes a mechanism for prioritizing data based on predicted connection times, which helps in reducing network congestion and ensuring timely data delivery. Similarly, \citet{dubey2016efficient} emphasizes the role of advanced scheduling techniques in minimizing latency and improving service quality. These approaches are crucial in dynamic environments where vehicular connections with RSUs are intermittent and unpredictable. Moreover, the allocation of resources in RSU-empowered vehicular networks is another critical area of focus. In \cite{tang2020intelligent}, the authors explore strategies to optimize data transmission and reduce latency through intelligent resource management. This study, along with \cite{ko2020rsu}, underscores the need for adaptive frameworks that can respond to changing network conditions and traffic patterns.

Ensuring reliable and efficient data transmission is essential for the effective operation of vehicular networks. \citet{lin2022multi} investigates the use of edge computing to offload data processing tasks, thereby enhancing throughput and reducing the computational burden on vehicles. Additionally, \citet{ali2016efficient} introduces network coding techniques to improve data dissemination, reduce redundancy, and optimize bandwidth usage. These studies highlight innovative solutions to common challenges in data-intensive vehicular networks. Moreover, the dynamic nature of vehicular networks necessitates robust frameworks for real-time data access and management. The research \cite{liu2010rsu} focuses on differentiating between safety-critical and non-safety-critical data, proposing a framework that adapts to traffic conditions to ensure timely access to vital information. This approach is critical for applications requiring immediate response, such as collision avoidance systems and real-time traffic updates. Lastly, integrating V2V (Vehicle-to-Vehicle) and V2I communication is explored in \cite{gao2020v2vr}. This study examines the potential of hybrid communication networks to enhance data transmission reliability and improve routing efficiency in complex urban settings. By leveraging both V2V and V2I communications, these networks can offer more resilient and flexible data transmission solutions.

These studies collectively emphasize the crucial role of RSUs in the broader context of vehicular communication networks. By addressing challenges related to data scheduling, resource allocation, data throughput, and real-time access, they provide a foundation for developing more efficient and reliable transportation systems. These advancements are essential for the successful implementation of CAVs, where seamless communication between vehicles and infrastructure is a cornerstone. However, the existing literature primarily focuses on developing and testing communication protocols, resource allocation strategies, and data dissemination techniques. Notably, there is a gap where real-world RSU data is integrated into simulations to generate diverse and high-fidelity traffic scenarios. Utilizing real-world RSU data could provide valuable insights into interactive vehicle behavior, traffic dynamics, and system performance under various conditions, thereby enhancing the fidelity and applicability of CAV simulations. The approach proposed in this study aims at bridging this gap between theoretical research and practical implementation, offering a more comprehensive understanding of the challenges and opportunities in deploying V2X technologies in real-world settings.

Specifically, the main contribution of this paper is threefold:

\begin{itemize}
\item This study processes and incorporates real-world Signal Phase and Timing (SPaT) data from RSUs into a simulation environment, which enhances its fidelity and applicability.

\item This study develops an algorithm to simulate how Autonomous Vehicles (AVs) operate and respond upon receiving real-time SPaT data through V2X communication, demonstrating practical V2I connectivity in the simulation environment.

\item The research generates a variety of traffic scenarios from vehicle interactions with real-world SPaT data, providing a comprehensive testing ground for CAVs.
\end{itemize}

The remainder of this paper is organized as follows. The second section details the methods used in this study. The third section presents the parameter and simulation setups and showcases the generated scenarios. The last section concludes the paper and provides future work.  

\section{Methodology}

\subsection{Framework}

\begin{figure}[!ht]
  \centering
  \includegraphics[width=0.9\textwidth]{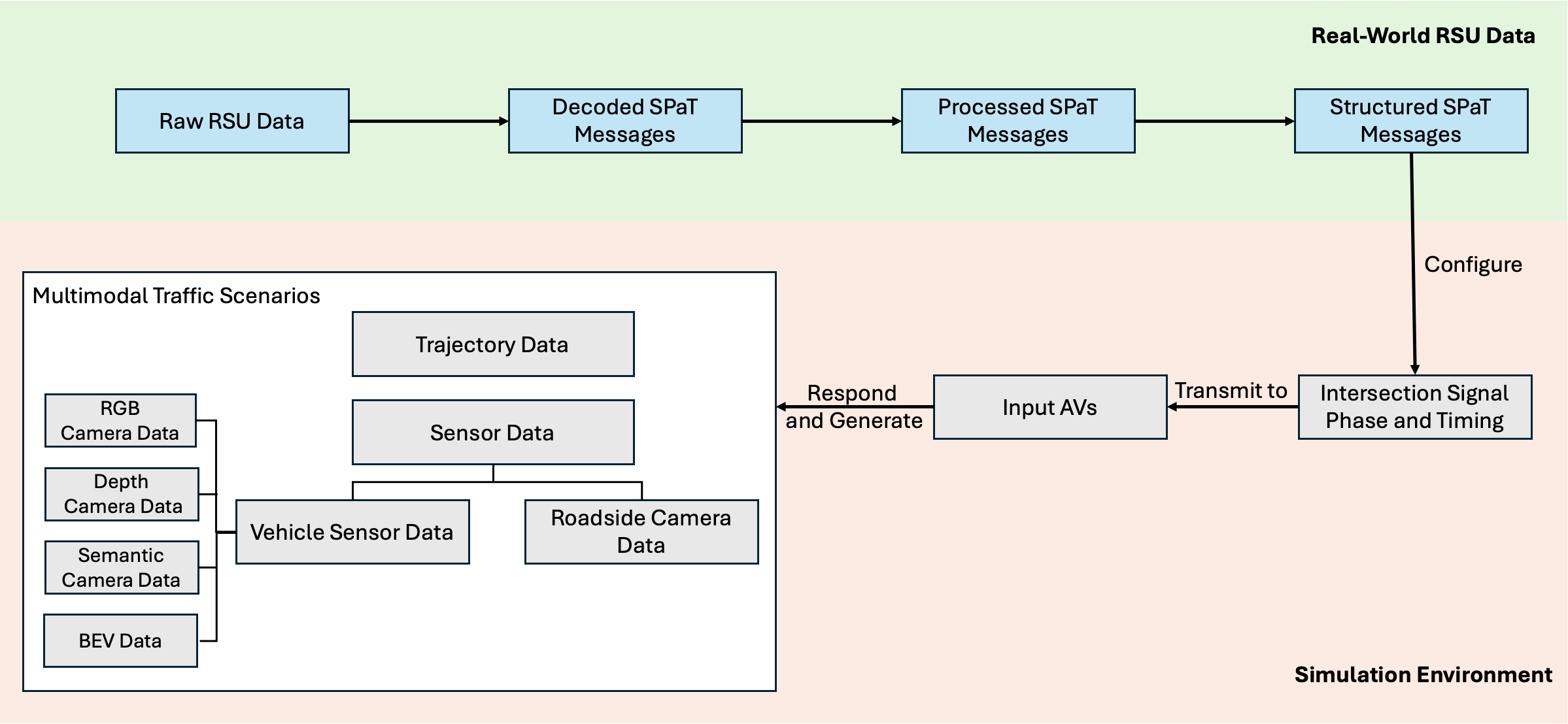}
  \caption{The Framework of The Interactive Traffic Scenario Generation Inspired by Real-World RSU Data}\label{fig:trial}
  \label{fig_1}
\end{figure}

The framework of this study is shown in Figure~\ref{fig_1}, which illustrates the process of integrating real-world RSU SPaT data into a simulation environment to generate comprehensive traffic scenarios. The upper part of the diagram shows the procedure for processing raw RSU data into structured SPaT messages. These messages provide detailed traffic signal timing information that is critical for a high-fidelity simulation environment. In the lower part of the diagram, the simulation environment is configured using the structured SPaT messages to accurately represent intersection signal phases and timings. AVs in the simulation can respond proactively to these real-time signal inputs once within the intersection area, simulating V2I interactions. This setup further generates multimodal traffic scenarios with detailed sensor data, including trajectory, RGB camera, and depth camera data from both vehicles and roadside sensors. The following sections will elaborate on each module in detail.

\subsection{Data Collection and Parsing}
\begin{figure}[!ht]
  \centering
  \includegraphics[width=0.4\textwidth]{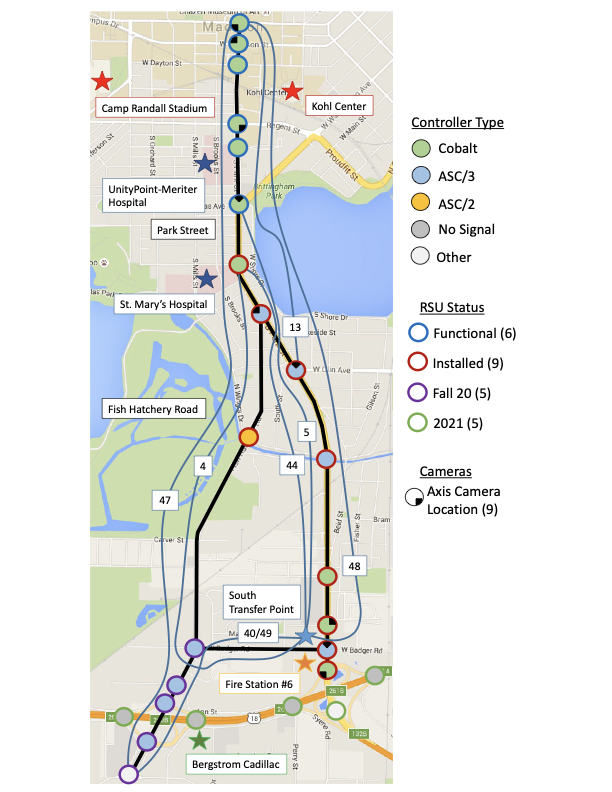}
  \caption{Layout of City of Madison Smart Corridor and RSU Distribution \cite{wu2023development}}\label{fig:trial}
  \label{fig_2}
\end{figure}

In general, encoded raw data is received distributively in three edge servers from RSUs placed on the existing arterial and freeway in Madison, WI, as shown in Figure~\ref{fig_2}. So far, 21 RSUs have been installed, including 15 RSUs installed along the Park Street and 6 RSUs installed along the Beltline freeway \cite{wu2023development, 10292930}.

\begin{figure}[!ht]
  \centering
  \includegraphics[width=0.7\textwidth]{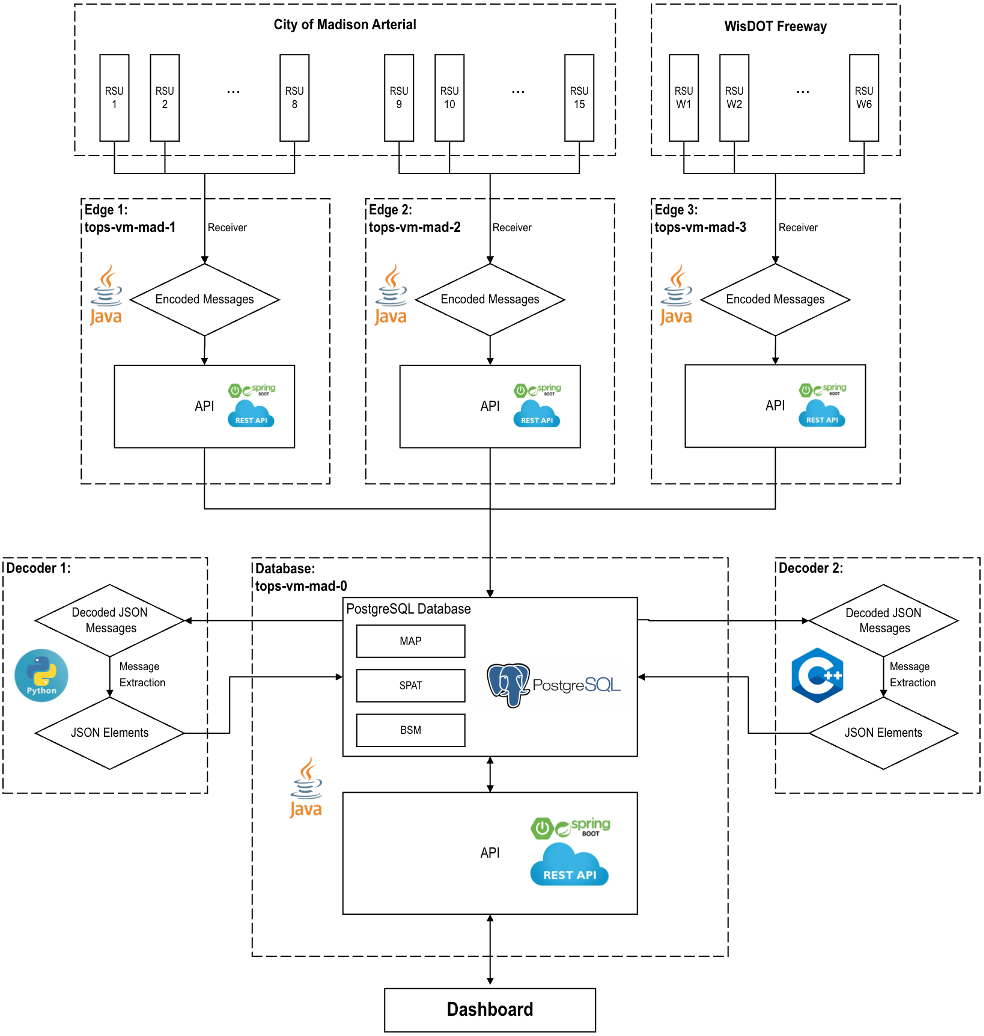}
  \caption{Data Pipeline of Smart Corridor}\label{fig:trial}
  \label{fig_3}
\end{figure}

\begin{figure}[!ht]
  \centering
  \includegraphics[width=0.35\textwidth]{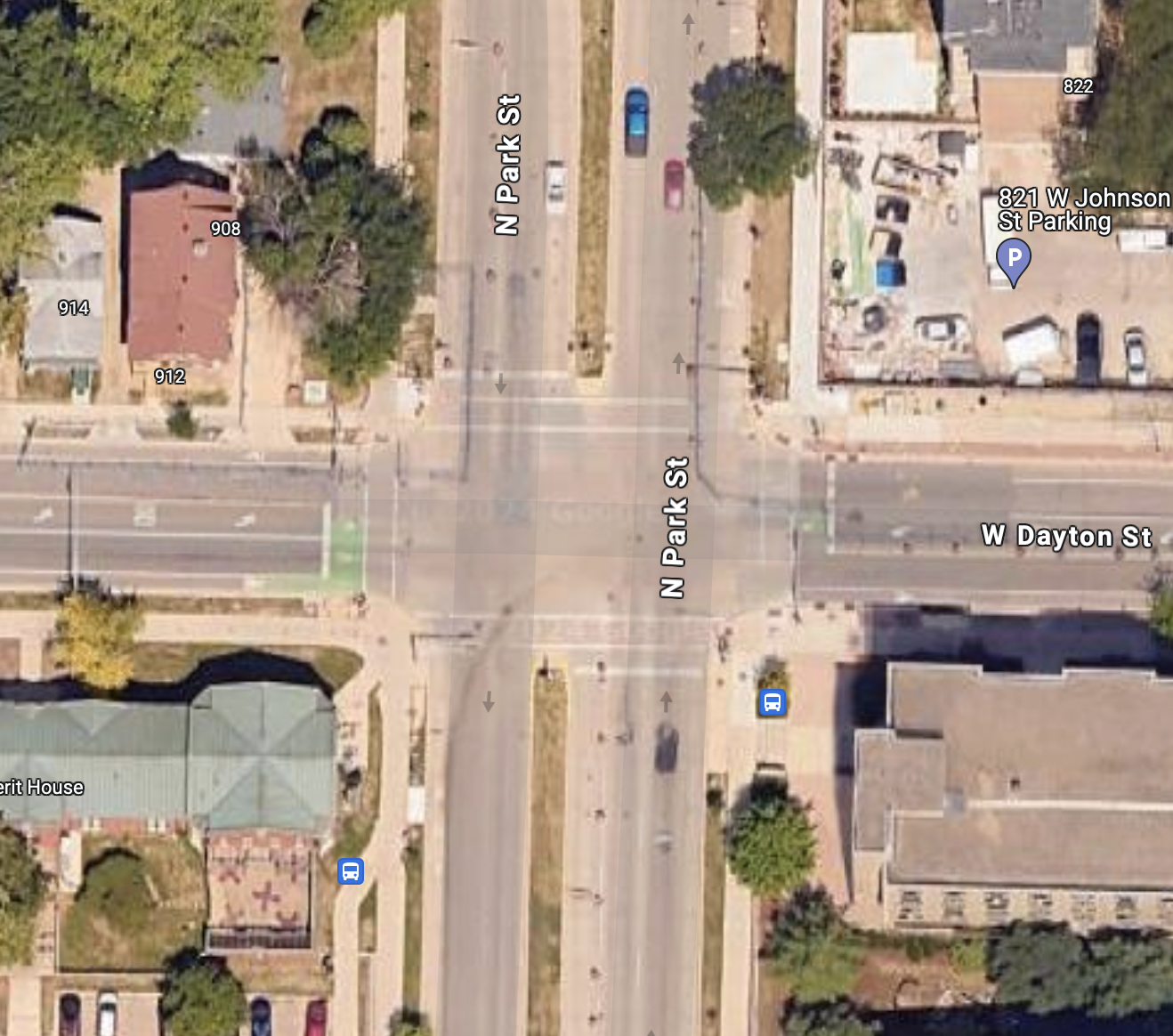}
  \caption{Intersection of Part Street and Dayton Street in Madison From Google Map}\label{fig:trial}
  \label{fig_4}
\end{figure}

\begin{figure}[!ht]
  \centering
  \includegraphics[width=0.8\textwidth]{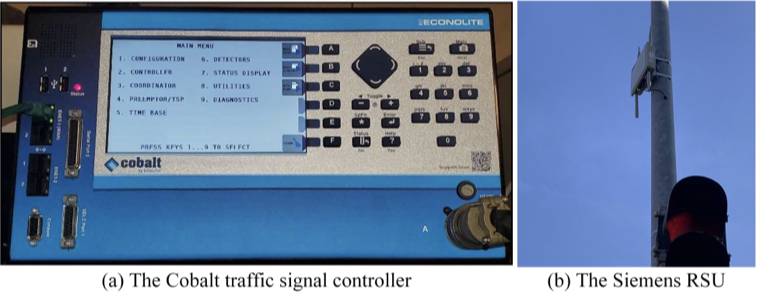}
  \caption{Signal Controller and RSU \cite{10292930}}\label{fig:trial}
  \label{fig_5}
\end{figure}

The received SPaT, MAP, and Basic Safety Messages (BSM), are stored in the database through API in another edge server. The database is connected with decoders for message decoding. The decoded messages are archived back in the database. The database is also connected to the SPaT dashboard through API for data display. The entire framework is shown in Figure~\ref{fig_3}. 
Specifically, this study aims to leverage SPaT data from the intersection of Part Street and Dayton Street, as shown in Figure~\ref{fig_4}, for interactive traffic scenario generation. The SPaT data is collected by the controller of the intersection and then transmitted to RSU. Figure ~\ref{fig_5} shows the signal controller and RSU used in this study. Utilizing a developed data pipeline, the encoded SPaT messages are then received from RSU and stored in the local database \cite{wu2023development}. 

The decoder used in this study to decode raw SPaT messages is the Pycrate library \cite{timmurphy.org2}, which is identified as an open-source toolkit for handling various telecommunications and networking protocols, such as  LTE (4G) and 5G NR. The library is fully licensed under LGPL v2.1, offering robust ASN.1 support for data description. Table 1 shows an example of a raw SPaT message and the corresponding decoded message using the Pycrate decoder. 

\begin{table}[!ht]
    \caption{Example of Raw SPaT Message and Decoded SPaT Message}\label{tab:versions}
    \begin{center}
        \small
        \begin{tabular}{>{\raggedright\arraybackslash}p{3cm} >{\raggedright\arraybackslash}p{14cm}}
            \hline
            \textbf{Raw Message} & \{"msg-wave":[\{"dot3":\{"chan":"SCH1","dest":"ffffffffffff","ll":3, \\ 
            & "priority":3,"psid":"8002","security":\{"cert":true,"crypt":false,"prof":false, \\ 
            & "sign":false\},"slot":"CONTINUOUS","xtension":0\},"encoding":"UPER", \\ 
            & "payload":"00133a44414b00863057f00008ab40700804302f498038218178940081180 \\
            & bbe600208a05df300304302f12802021817a4c0141140bbe600c08c05df30"\}], \\ 
            & "seqno":1, "pkgno":0\} \\
            \hline
            \textbf{Payload} & 00133a44414b00863057f00008ab40700804302f498038218178940081180 \\
            & bbe600208a05df300304302f12802021817a4c0141140bbe600c08c05df30 \\ 
            \hline
            \textbf{Decoded Message} & \{"messageId":19,"value":\{"timeStamp":278859,"intersections": \\
            & [\{"timeStamp":35508,"id":\{"id":50698\},"revision":127,"status":"0000","states": \\
            & [\{"state-time-speed":[\{"eventState":"stop-And-Remain","timing"\:{"minEndTime":24211}\}], \\
            & "signalGroup":8\},\{"state-time-speed":[\{"eventState":"stop-And-Remain","timing": \\
            & \{"minEndTime":24101\}\}],"signalGroup":7\},\{"state-time-speed": \\
            & [\{"eventState":"protected-Movement-Allowed","timing":\{"minEndTime":24051\}\}], \\
            & "signalGroup":2\},\{"state-time-speed":[\{"eventState":"permissive-Movement-Allowed", \\
            & "timing":\{"minEndTime":24051\}\}],"signalGroup":1\},\{"state-time-speed": \\ 
            & [\{"eventState":"stop-And-Remain","timing":\{"minEndTime":24101\}\}], \\
            & "signalGroup":3\},\{"state-time-speed":[\{"eventState":"stop-And-Remain", \\
            & "timing":\{"minEndTime":24211\}\}],"signalGroup":4\},\{"state-time-speed":[\{"eventState": \\
            & "permissive-Movement-Allowed","timing":\{"minEndTime":24051\}\}],"signalGroup":5\}, \\
            & \{"state-time-speed":[\{"eventState":"protected-Movement-Allowed","timing": \\
            & \{"minEndTime":24051\}\}],"signalGroup":6\}]\}]\}\} \\
            \hline
        \end{tabular}
    \end{center}
\end{table}

The decoded message adheres to the SAE J2735 standard for Dedicated Short Range Communications (DSRC) in vehicular environments. It contains a variety of fields, starting with the "messageId," which identifies the type of message, likely a SPaT message used to convey traffic signal states. The "value" field encompasses the main content of the message, including the "timeStamp" that indicates when the message was generated, ensuring synchronization across systems. Within the "intersections" section, details about one or more intersections are provided, including a unique "id" for each intersection, ensuring accurate identification. The "revision" field denotes the version or update status of the intersection data, maintaining the currency of information. The "status" field may reflect the operational state of the intersection, indicating normal operation or highlighting issues. The "states" array lists the current states of traffic signals at the intersection, describing each with specific attributes. The "eventState" indicates the current signal phase, such as "Stop-And-Remain" or "protected-Movement-Allowed," dictating the actions permitted for vehicles. The "timing" information, including fields like "minEndTime," provides details on the timing of these states, such as the minimum remaining time for the current signal phase. The "signalGroup" identifies groups of signals controlled together, typically representing sets of lanes or directions at the intersection. This standardized format ensures consistent communication between vehicles and infrastructure effectively.

The decoded SPaT messages are further processed clearly for easier interpretation. The processed SPaT message of the above example is illustrated in Table 2. This table presents a processed and readable format of a SPaT message for the intersection noted as "Dayton", where "Timestamp" indicates the time the message was generated. The table details various signal groups, numbered 1 through 8, each associated with specific lanes or directions at the intersection. For each signal group, it provides an event state, such as 'permissive-Movement-Allowed' or 'stop-And-Remain,' describing the current traffic signal phase. Additionally, it lists the remaining time in seconds for each signal state currently. The processed SPaT messages are utilized to configure the signal phases of an intersection in the simulation environment.

\begin{table}[!ht]
	\caption{Example of Processed Decoded SPaT Message}\label{tab:versions}
	\begin{center}
		\begin{tabular}{l l}
			\textbf{Intersection} & Dayton \\\hline
                \textbf{Timestamp} & 12:39:36 AM \\\hline
                Signal Group 1 & ['permissive-Movement-Allowed', 29.592]  \\
                Signal Group 2 & ['protected-Movement-Allowed', 29.592] \\
                Signal Group 3 & ['stop-And-Remain', 34.592]  \\
                Signal Group 4 & ['stop-And-Remain', 45.592]  \\
                Signal Group 5 & ['permissive-Movement-Allowed', 29.592]  \\
                Signal Group 6 & ['protected-Movement-Allowed', 29.592]  \\
                Signal Group 7 & ['stop-And-Remain', 34.592]  \\
                Signal Group 8 & ['stop-And-Remain', 45.592]  \\\hline
		\end{tabular}
	\end{center}
\end{table}

\subsection{V2X Collaborative Autonomous Driving Design}
To ensure fully connected and automated traffic, where AVs are connected with infrastructure through V2I communication towards collaborative autonomous driving, in the CARLA simulator \cite{Dosovitskiy17}, we need to carefully design the logic for vehicle operation. While approaching an intersection to a certain distance, AVs need to in advance gain access to the signal time remaining in the specific direction through which they are driving, and respond correspondingly. 

By proactively obtaining signal timing information from the infrastructure, AVs can optimize their speed and trajectory well in advance, leading to smoother and more efficient navigation through intersections. This reduces the likelihood of abrupt braking or acceleration, which not only enhances passenger comfort but also improves overall traffic flow and reduces energy consumption. Furthermore, this strategy minimizes the risk of collisions by providing vehicles with accurate and timely information, enabling them to make informed decisions and avoid potential hazards. While high-level AVs with advanced sensing systems can detect and interpret traffic signals, they highly rely on vehicle sensors, which can be affected by various environmental factors such as poor lighting conditions, adverse weather, and obstructions. These factors can compromise the accuracy and reliability of the information obtained. In contrast, direct transmission of signal timing information via V2I communication ensures that AVs receive precise and real-time data, irrespective of external conditions. This method not only enhances the reliability of the information but also reduces the computational burden on the vehicles' onboard systems, allowing them to focus on other critical tasks. Hence, in this study, we aim to design and simulate the behavior of AVs in the CARLA simulator as they receive real-time traffic signal data upon entering a certain range. This simulation will enable us to generate high-fidelity and complex multimodal traffic scenarios, providing a rich dataset for further research. 

\begin{algorithm}
\footnotesize 
\caption{Proactive Intersection Navigation for AVs}
\begin{algorithmic}[1]
\WHILE{simulation is running}
    \STATE $L_v \gets$ \text{current location of} \textit{vehicle}
    \IF{$L_v$ \text{is within the intersection area}}
        \STATE $v_c \gets$ \text{current speed of} \textit{vehicle}
        \STATE $W_v \gets$ \text{location of} \textit{vehicle} \text{on the map}
        \STATE $T_t \gets$ \text{identify target traffic light for} $W_v$ \text{from} \textit{controlled\_traffic\_lights}
        
        \IF{$T_t$ \text{is not} \texttt{None}}
            \STATE $S_t \gets$ \text{state of} $T_t$
            \STATE $D_t \gets$ \text{distance from} $L_v$ \text{to} $T_t$
            \STATE $G_r \gets$ \text{remaining green light duration of} $T_t$
            
            \IF{$S_t$ \text{is} \texttt{Red}}
                \IF{$D_t < 10$ \text{meters}}
                    \STATE $v_t \gets 0.0$
                \ELSE
                    \STATE $v_t \gets 0.5 \times v_c$
                \ENDIF
            \ELSIF{$S_t$ \text{is} \texttt{Yellow}}
                \IF{$D_t < 10$ \text{meters}}
                    \STATE $v_t \gets 0.0$
                \ELSE
                    \STATE $v_t \gets 0.5 \times v_c$
                \ENDIF
            \ELSIF{$S_t$ \text{is} \texttt{Green}}
                \STATE $v_t \gets$ \text{SPEED\_LIMIT}
            \ENDIF
            
            \STATE $\mathbf{v}_f \gets$ \text{forward vector of} \textit{vehicle}
            \STATE $\mathbf{v}_t \gets \mathbf{v}_f \times v_t$
            \STATE \text{Set target speed of} \textit{vehicle} \text{to} $\mathbf{v}_t$
        \ENDIF
    \ENDIF
    \STATE \text{Wait for 0.1 seconds}
\ENDWHILE
\end{algorithmic}
\end{algorithm}

Algorithm 1 demonstrates in detail how AVs respond and operate upon receiving real-time traffic signal data. AVs are enabled in CARLA by applying the Autopilot function in advance. In the CARLA simulator, the Autopilot feature refers to a built-in, AI-controlled driving system that allows vehicles to navigate and operate autonomously within the simulated environment. Specifically, the proposed algorithm initiates by continuously monitoring the AV's location and determining if it is within a predetermined intersection area. Upon entering this zone, the AV's current speed and location are assessed, and the relevant traffic light for its direction is identified. 
Since AVs receive traffic signal information in advance, they can respond proactively to navigate the intersection. Based on the traffic light state—red, yellow, or green—the AV's target speed is adjusted preemptively. For instance, if the light is red and the AV is close to the intersection, it will come to a stop. If the light is yellow, the AV will slow down significantly. If the light is green, the AV will proceed at the speed limit. Since the Autopilot function is already enabled, AVs have been equipped to handle complex conditions. This approach simplifies the AV control strategy by focusing on connectivity, allowing vehicles to react in advance to the received signal information, thus managing intersection navigation effectively.

\section{Experiment}
With the proactive intersection navigation algorithm in place, we can simulate AVs in CARLA using real-world signal data. Specifically, we utilize the processed SPaT data from the intersection of Park Street and Dayton Street, collected, decoded, and processed as described earlier. We further structure the data into a sequence of signal phases derived from consecutive processed SPaT messages. Table 3 below illustrates a representative sample of the traffic light phases used in the simulation, which details the duration and state of traffic lights for both directions during each phase. The structured data is eligible to be used to configure the signal phase and timing of a four-way intersection in CARLA. 

\begin{table}[!ht]
    \centering
    \caption{Traffic Light Phases Structured from SPaT Data}
    \begin{tabular}{cccc}
        \hline
        \textbf{Phase} & \textbf{Duration (Seconds)} & \textbf{North-South Direction} & \textbf{East-West Direction} \\ \hline
        1 & 28 & Green & Red \\ 
        2 & 3 & Yellow & Red \\ 
        3 & 20 & Red & Green \\ 
        4 & 3 & Red & Yellow \\ \hline
        1 & 35 & Green & Red \\ 
        2 & 3 & Yellow & Red \\ 
        3 & 20 & Red & Green \\ 
        4 & 3 & Red & Yellow \\ \hline
        1 & 35 & Green & Red \\ 
        2 & 3 & Yellow & Red \\ 
        3 & 20 & Red & Green \\ 
        4 & 3 & Red & Yellow \\ \hline
    \end{tabular}
    \label{tab:traffic_phases}
\end{table}

\subsection{Simulation Setup}
We use the default map Town 10 in Carla as shown in Figure~\ref{fig_6} (a), where the target four-way intersection is marked by the yellow circle. Figure~\ref{fig_6} (b) marks the traffic signals of this intersection. he signal phase and timing of this intersection are set using the real-world SPaT data, as described in the previous section. The number of vehicles spawned is 100, and the speed limit is set to 15 $m/s$. The intersection area in this study is defined as a square extending 35 meters in each direction from the center of the intersection. The center of the intersection $\text{C}_{\text{intersection}}$ can be calculated based on the exact locations of the traffic signals, as shown in the following equation:

\begin{figure}[!ht]
  \centering
  \includegraphics[width=1.0\textwidth]{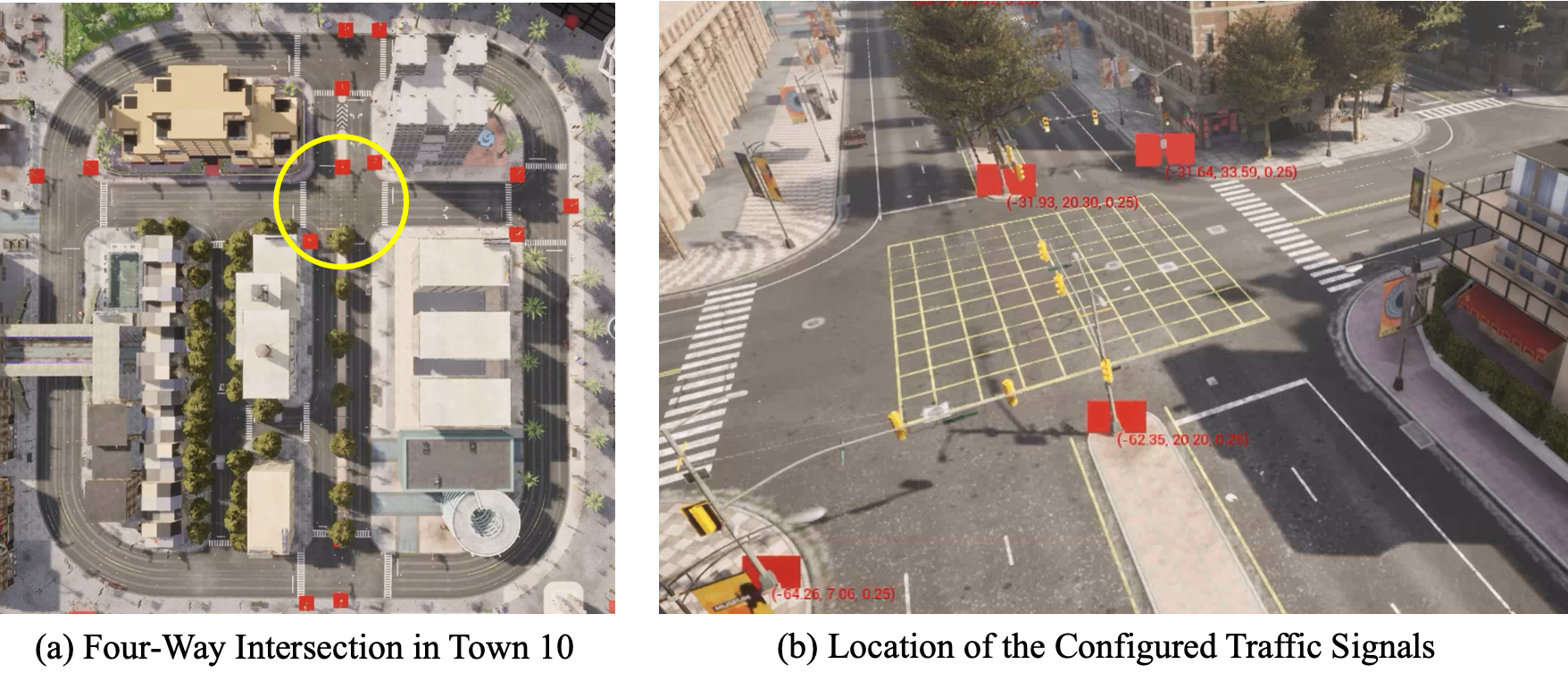}
  \caption{Simulation Intersection and Signals}\label{fig:trial}
  \label{fig_6}
\end{figure}

\begin{linenomath}
  \begin{equation}
  \text{C}_{\text{intersection}} = \frac{\sum_{i=1}^{n} \text{NS}_i + \sum_{j=1}^{m} \text{EW}_j}{n + m}
  \end{equation}
\end{linenomath}

where $\text{NS}_i$ represents the location of the $i$-th traffic light in the north-south direction, $\text{EW}_i$ represents the location of the $j$-th traffic light in the east-west direction, $n$ is the total number of traffic lights in the north-south direction, and $m$ is the total number of traffic lights in the east-west direction. Normally, $m = n =2$. This formula calculates the centroid of the traffic light locations. The simulation begins with the start of the first cycle and terminates at the end of the last cycle.

\subsection{Interactive Scenario Generation}
The objective of the simulation is to generate interactive traffic scenarios, which involve real-time interactions between AVs and other road users and the infrastructure. In this study, the interaction is uniquely facilitated and enhanced by the communication of real-world traffic signal data, which allows AVs to adjust their behavior proactively and dynamically based on traffic conditions.

\begin{figure}[!ht]
  \centering
  \includegraphics[width=0.8\textwidth]{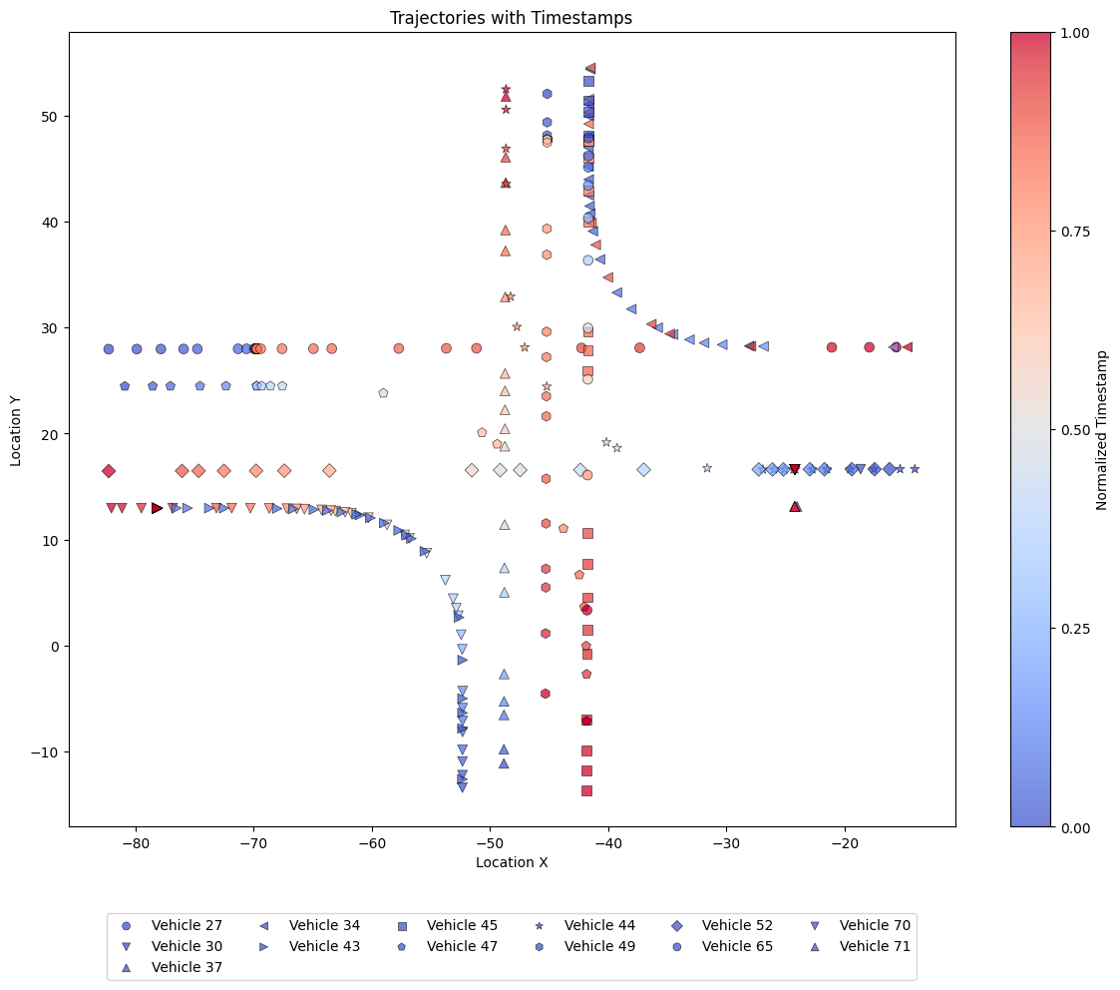}
  \caption{Simulation Intersection and Signals}\label{fig:trial}
  \label{fig_7}
\end{figure}

In these scenarios, two main types of data are collected: trajectory data and sensor data. The trajectory data includes detailed records of the AVs' movements, such as position, speed, and acceleration. Figure~\ref{fig_7} illustrates the movement patterns of various vehicles entering and passing through the intersection area. It shows the vehicle position in a two-dimensional plane. The color gradient indicates the temporal progression of the vehicle trajectories, as well as the sequence in which different vehicles arrive at and pass through the intersection. These trajectory are the direct result of AVs responding to the received real-time traffic signal data.

\begin{figure}[!ht]
  \centering
  \includegraphics[width=1\textwidth]{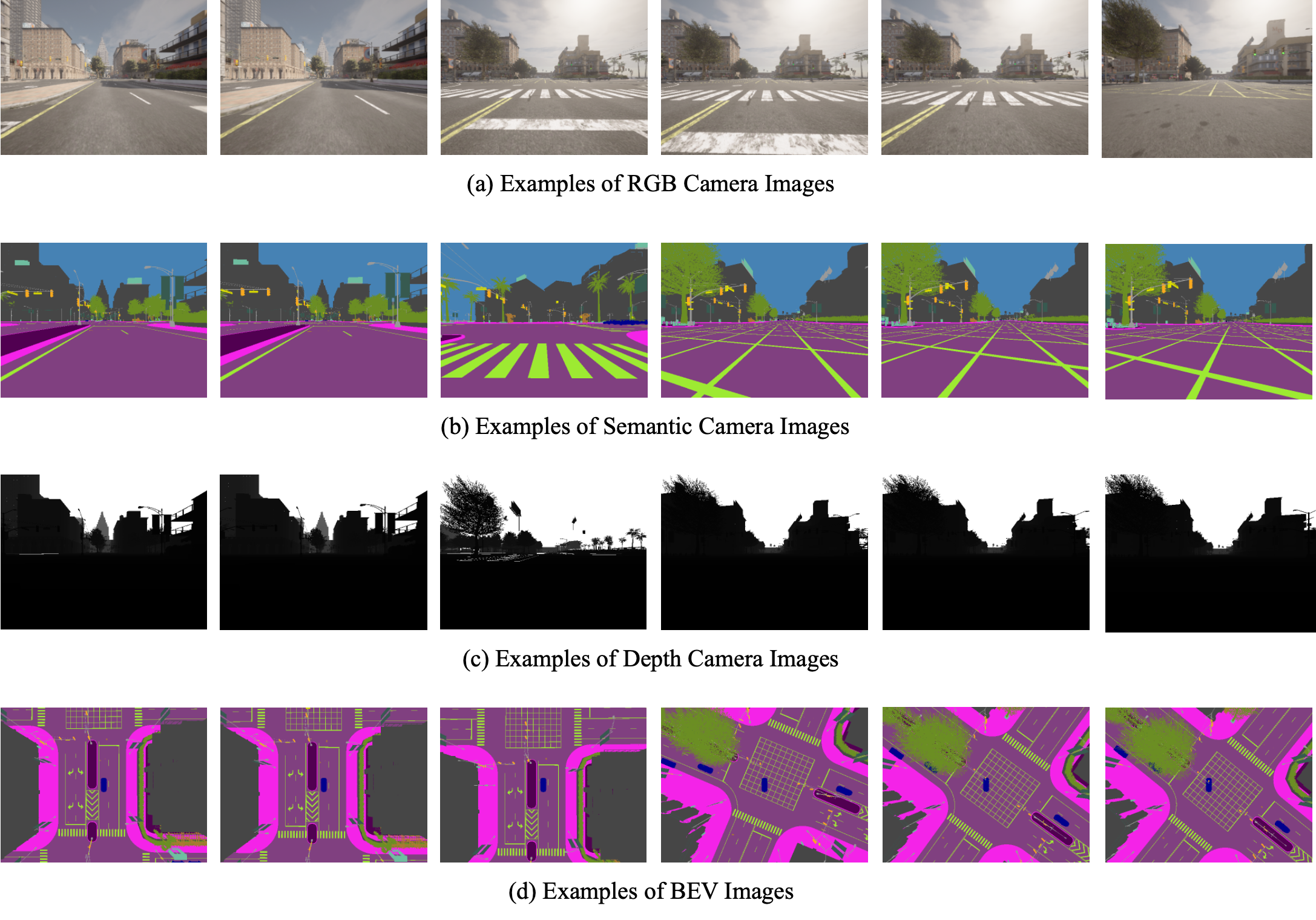}
  \caption{Various Types of Vehicle Sensor Data}\label{fig:trial}
  \label{fig_8}
\end{figure}

Sensor data encompasses both vehicle and infrastructure sensor data, each offering critical insights into the environment around the AV. Vehicle sensor data, gathered from the onboard cameras, provides a detailed account of the vehicle's immediate environment, including other vehicles, static obstacles, drivable areas, lane markings, and traffic signals. This data is essential for real-time navigation and decision-making. Additionally, roadside camera data complements this information by offering a broader perspective of the traffic scenario, enhancing the overall situational awareness of the AV system. Figure~\ref{fig_8} illustrates various types of sensor data captured during the simulation, showcasing the different imaging modalities used to gather comprehensive environmental information.
The cameras used in these simulations are carefully configured to optimize the quality and relevance of the data collected, as listed below:

\begin{itemize}
\item RGB Camera Images in Figure~\ref{fig_8} (a): These images provide a standard visual representation of the environment, showing the appearance of surroundings, which may include vehicles, lane marks, road signs, and other objects. The RGB camera is configured with a resolution of 640 x 480 pixels and a 110-degree field of view (FOV), capturing detailed visual data fundamental for tasks such as object detection and recognition.

\item Semantic Camera Images in Figure~\ref{fig_8} (b): These images are processed to segment the scene into different categories, which highlighs key features such as lanes, drivable areas, and various object types. The semantic camera uses the same resolution and FOV as the RGB camera. Different colors in these images represent various objects and areas, aiding the AV in understanding the structure of the environment and assisting in tasks like lane keeping and navigation.

\item Depth Camera Images in Figure~\ref{fig_8} (c): These images provide information on the distance between the vehicle and surrounding objects, represented in shades of gray. Darker tones indicate closer objects, while lighter tones denote further distances. The depth camera, configured with the same 640 x 480 pixel resolution and 110-degree FOV, is crucial for spatial awareness and collision avoidance, which helps the vehicle gauge distances accurately.

\item Bird's Eye View (BEV) Images in Figure~\ref{fig_8} (d): BEV images offer an overhead perspective, providing a comprehensive layout of the intersection and surrounding areas. These images are captured 25 meters above the vehicle with a pitch angle of -90 degrees, offering a top-down view essential for monitoring traffic flow, vehicle positioning, and strategic planning in dense traffic situations. The also camera uses the same resolution and FOV to ensuring consistency in data quality across different views.
\end{itemize}

These rich vehicle sensor data, acting as an indirect viewpoint of the AVs' response to traffic signal data, offer insights into the vehicle's status from an environmental perspective. This data is crucial for understanding how the connected and automated vehicle's intrinsic autonomous driving capabilities harmonize with external information from connected infrastructure, highlighting the integration and synergy between onboard systems and external guidance.

\begin{figure}[!ht]
  \centering
  \includegraphics[width=1\textwidth]{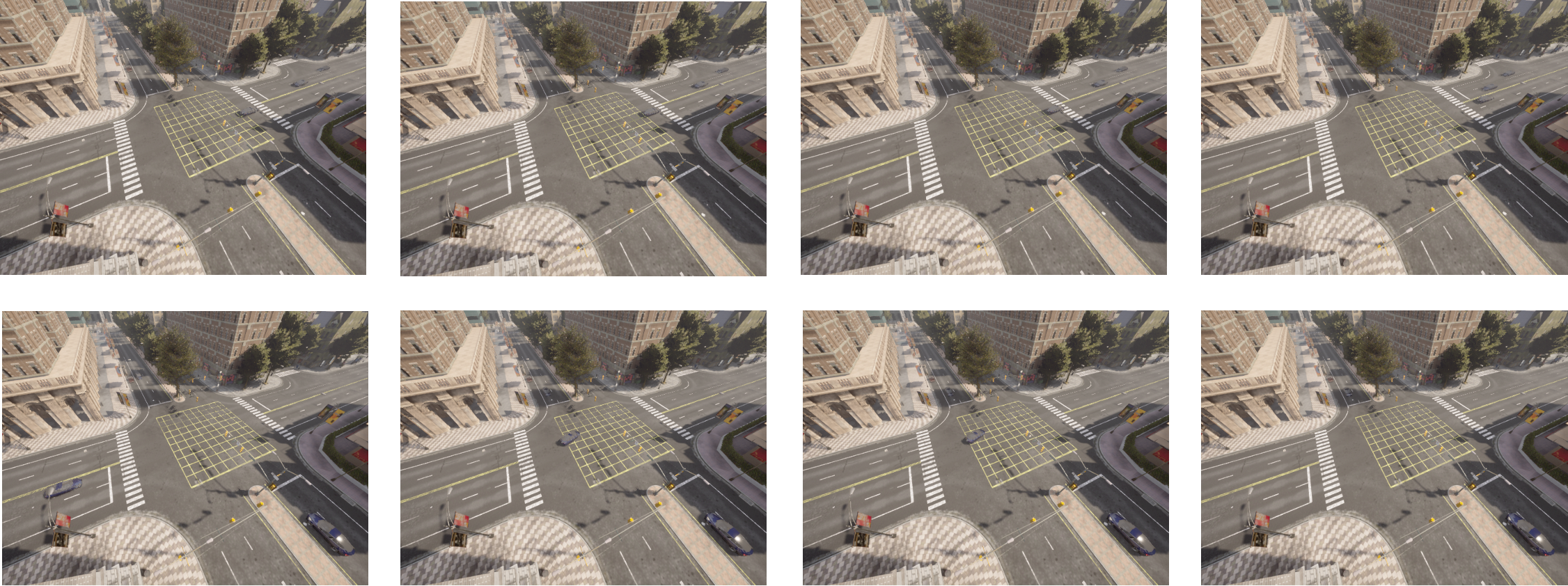}
  \caption{Example of Images Captured by Roadside Camera}\label{fig:trial}
  \label{fig_9}
\end{figure}

Roadside camera data, as shown in Figure~\ref{fig_9}, offers a broader view of the intersection, capturing the overall traffic flow and providing an external validation of the AVs' actions. This data is invaluable for cross-verifying the accuracy and effectiveness of the vehicle's responses to the traffic signals and the actions of other road users. The roadside camera is configured to capture high-quality visual data with an RGB camera set to a resolution of 640 x 480 pixels and a 110-degree FOV. It is strategically positioned at a height of 20 meters above the ground, with a pitch angle of -50 degrees and a yaw of 25 degrees, which provides a comprehensive and angled view of the intersection area. Moreover, Table 4 summarizes the types of sensors used in the simulation, along with their specific parameters and unique features. These configurations are designed to be flexible and versatile, allowing adjustments based on personalized user needs and specific research objectives. This adaptability ensures that the data captured is highly relevant and useful for analyzing the AVs' behaviors and decision-making processes in response to real-time traffic conditions.

\begin{table}[!ht]
    \centering
    \caption{Summary of Sensors and Configurations}
    \begin{tabular}{cccc}
        \hline
        \textbf{Sensor Type} & \textbf{Resolution} & \textbf{FOV} & \textbf{Additional Configuration} \\ \hline
        RGB Camera & 640 x 480 & 110° & Standard visual rep. \\ 
        Semantic Camera & 640 x 480 & 110° & Scene segmentation \\ 
        Depth Camera & 640 x 480 & 110° & Distance measurement \\ 
        BEV Camera & 640 x 480 & 110° & Overhead, 25 m height, -90° pitch \\ 
        Roadside Camera & 640 x 480 & 110° & 20 m height, -50° pitch, 25° yaw \\ \hline
    \end{tabular}
    \label{tab:sensor_summary}
\end{table}

The integration of trajectory and sensor data provides a robust foundation for a comprehensive analysis of the AVs' performance, especially regarding safety, efficiency, and compliance with traffic regulations. Utilizing real-world traffic signal data allows researchers to observe and evaluate the AVs' behavior under real traffic conditions, highlighting the vehicles' responses to dynamic signal changes and the behavior of other road users. The scenarios generated in this experiment, underpinned by real-world signal data, serve as an essential testing ground for the advancement of CAV technologies. They provide invaluable insights into how AVs can be more effectively integrated with existing traffic management systems, and potentially foster the development of more intelligent and adaptive traffic solutions. The use of real-world data not only enhances the fidelity, realism and reliability of the simulations but also ensures that the findings apply to realistic urban settings. 

\section{Conclusion}
This study underscores the significance of incorporating real-world traffic data into simulation frameworks to advance the evaluation and implementation of CAVs. By integrating real-world SPaT data from RSUs into the CARLA simulation environment, we have created a high-fidelity and applicable testing ground for evaluating V2X communication utilized by AVs. The key contributions of this work include developing an algorithm that simulates AVs' response to real-time traffic signals and generating comprehensive traffic scenarios. These efforts provide a robust framework for analyzing the effectiveness and reliability of V2X communication systems, highlighting the interactions between AVs and other road users and the infrastructure. The use of real-world data enhances the fidelity of the simulation. Moreover, the framework offers a variety of sensor outputs in dynamic traffic scenarios, providing comprehensive data for analyzing AV behavior, traffic flow, safety dynamics, etc. Such insights are vital for refining AV algorithms with enhanced safety and efficiency.

Future research should aim to incorporate additional real-world data, including data related to environmental conditions, pedestrian activities, and unexpected traffic events, to further enrich the simulation environment. For example, expanding the simulation environments from urban to rural conditions will provide a more comprehensive understanding of V2X system performance across diverse contexts. Furthermore, the development of advanced predictive models could enhance the responsiveness and efficiency of AVs in dynamic traffic scenarios, optimizing their interactions with both infrastructure and other road users.

\section{Acknowledgements}
The Park Street Smart Corridor is being developed through a collaboration of the TOPS Lab, the City of Madison, Traffic and Parking Control Products and Solutions (TAPCO), and the Wisconsin Department of Transportation.  The ideas and views expressed in this paper are strictly those of the TOPS Lab a the University of Wisconsin.

\section{Author Contributions}
The authors confirm their contribution to the paper as follows: study conception and design: Junwei You, Pei Li, Yang Cheng, Keshu Wu, Rui Gan, Steven T. Parker; algorithm development and program development: Junwei You, Pei Li, Yang Cheng, Keshu Wu, Rui Gan; data preparation and analysis: Junwei You, Pei Li, Keshu Wu; manuscript preparation: Junwei You, Pei Li, Yang Cheng, Keshu Wu, Rui Gan, Steven T. Parker, Bin Ran. All authors reviewed the results and approved the final version of the manuscript.

\newpage

\bibliographystyle{trb}
\bibliography{trb_template}

\begin{thebibliography}{25}
\providecommand{\natexlab}[1]{#1}

\bibitem[{Huang et~al.(2024)Huang, Chen, Pian, Sheng, Ahn, and Noyce}]{huang2024toward}
Huang, Z., S.~Chen, Y.~Pian, Z.~Sheng, S.~Ahn, and D.~A. Noyce, Toward C-V2X Enabled Connected Transportation System: RSU-Based Cooperative Localization Framework for Autonomous Vehicles. \emph{IEEE Transactions on Intelligent Transportation Systems}, 2024.

\bibitem[{Hobert et~al.(2015)Hobert, Festag, Llatser, Altomare, Visintainer, and Kovacs}]{hobert2015enhancements}
Hobert, L., A.~Festag, I.~Llatser, L.~Altomare, F.~Visintainer, and A.~Kovacs, Enhancements of V2X communication in support of cooperative autonomous driving. \emph{IEEE communications magazine}, Vol.~53, No.~12, 2015, pp. 64--70.

\bibitem[{You et~al.(2024)You, Shi, Jiang, Huang, Gan, Wu, Cheng, Li, and Ran}]{you2024v2x}
You, J., H.~Shi, Z.~Jiang, Z.~Huang, R.~Gan, K.~Wu, X.~Cheng, X.~Li, and B.~Ran, V2X-VLM: End-to-End V2X Cooperative Autonomous Driving Through Large Vision-Language Models. \emph{arXiv preprint arXiv:2408.09251}, 2024.

\bibitem[{Fu et~al.(2021)Fu, Zhang, and Jiang}]{fu2021network}
Fu, S., W.~Zhang, and Z.~Jiang, A network-level connected autonomous driving evaluation platform implementing C-V2X technology. \emph{China Communications}, Vol.~18, No.~6, 2021, pp. 77--88.

\bibitem[{Choudhury et~al.(2016)Choudhury, Maszczyk, Asif, Mitrovic, Math, Li, and Dauwels}]{choudhury2016integrated}
Choudhury, A., T.~Maszczyk, M.~T. Asif, N.~Mitrovic, C.~B. Math, H.~Li, and J.~Dauwels, An integrated V2X simulator with applications in vehicle platooning. In \emph{2016 IEEE 19th International Conference on Intelligent Transportation Systems (ITSC)}, IEEE, 2016, pp. 1017--1022.

\bibitem[{Grimm et~al.(2024)Grimm, Schindewolf, Kraus, and Sax}]{grimm2024co}
Grimm, D., M.~Schindewolf, D.~Kraus, and E.~Sax, Co-simulate no more: The CARLA V2X Sensor. In \emph{2024 IEEE Intelligent Vehicles Symposium (IV)}, IEEE, 2024, pp. 2429--2436.

\bibitem[{Lee et~al.(2019)Lee, Wang, Wu, Kuo, Huang, Wang, Lin, Chen, and Tseng}]{lee2019building}
Lee, T.-K., T.-W. Wang, W.-X. Wu, Y.-C. Kuo, S.-H. Huang, G.-S. Wang, C.-Y. Lin, J.-J. Chen, and Y.-C. Tseng, Building a V2X simulation framework for future autonomous driving. In \emph{2019 20th Asia-Pacific Network Operations and Management Symposium (APNOMS)}, IEEE, 2019, pp. 1--6.

\bibitem[{Eckermann et~al.(2019)Eckermann, Kahlert, and Wietfeld}]{eckermann2019performance}
Eckermann, F., M.~Kahlert, and C.~Wietfeld, Performance analysis of C-V2X mode 4 communication introducing an open-source C-V2X simulator. In \emph{2019 IEEE 90th Vehicular Technology Conference (VTC2019-Fall)}, IEEE, 2019, pp. 1--5.

\bibitem[{Chen et~al.(2021)Chen, Chai, Hu, Jiang, and He}]{chen2021performance}
Chen, M., R.~Chai, H.~Hu, W.~Jiang, and L.~He, Performance evaluation of C-V2X mode 4 communications. In \emph{2021 IEEE Wireless Communications and Networking Conference (WCNC)}, IEEE, 2021, pp. 1--6.

\bibitem[{Zhang and Masoud(2020)}]{zhang2020v2xsim}
Zhang, E. and N.~Masoud, V2xsim: A v2x simulator for connected and automated vehicle environment simulation. In \emph{2020 IEEE 23rd International Conference on Intelligent Transportation Systems (ITSC)}, IEEE, 2020, pp. 1--6.

\bibitem[{Queck et~al.(2008)Queck, Sch{\"u}nemann, Radusch, and Meinel}]{queck2008realistic}
Queck, T., B.~Sch{\"u}nemann, I.~Radusch, and C.~Meinel, Realistic simulation of v2x communication scenarios. In \emph{2008 IEEE Asia-Pacific Services Computing Conference}, IEEE, 2008, pp. 1623--1627.

\bibitem[{Sch{\"u}nemann(2011)}]{schunemann2011v2x}
Sch{\"u}nemann, B., V2X simulation runtime infrastructure VSimRTI: An assessment tool to design smart traffic management systems. \emph{Computer Networks}, Vol.~55, No.~14, 2011, pp. 3189--3198.

\bibitem[{Qom et~al.(2016)Qom, Xiao, and Hadi}]{qom2016evaluation}
Qom, S.~F., Y.~Xiao, and M.~Hadi, Evaluation of cooperative adaptive cruise control (CACC) vehicles on managed lanes utilizing macroscopic and mesoscopic simulation. In \emph{Proceedings of the Transportation Research Board 95th Annual Meeting}, 2016, 16-6384.

\bibitem[{Mershad and Artail(2012)}]{mershad2012score}
Mershad, K. and H.~Artail, SCORE: Data scheduling at roadside units in vehicle ad hoc networks. In \emph{2012 19th International Conference on Telecommunications (ICT)}, IEEE, 2012, pp. 1--6.

\bibitem[{Dubey et~al.(2016)Dubey, Chauhan, and Chand}]{dubey2016efficient}
Dubey, B.~B., N.~Chauhan, and N.~Chand, Efficient data scheduling technique at RSU for vehicular ad-hoc networks. In \emph{2016 International Conference on Information Communication and Embedded Systems (ICICES)}, IEEE, 2016, pp. 1--7.

\bibitem[{Tang et~al.(2020)Tang, Zhu, Wei, Wu, Li, and Rodrigues}]{tang2020intelligent}
Tang, C., C.~Zhu, X.~Wei, H.~Wu, Q.~Li, and J.~J. Rodrigues, Intelligent resource allocation for utility optimization in RSU-empowered vehicular network. \emph{IEEE access}, Vol.~8, 2020, pp. 94453--94462.

\bibitem[{Ko et~al.(2020)Ko, Liu, Son, and Park}]{ko2020rsu}
Ko, B., K.~Liu, S.~H. Son, and K.-J. Park, RSU-assisted adaptive scheduling for vehicle-to-vehicle data sharing in bidirectional road scenarios. \emph{IEEE Transactions on Intelligent Transportation Systems}, Vol.~22, No.~2, 2020, pp. 977--989.

\bibitem[{Lin et~al.(2022)Lin, Huang, and Wu}]{lin2022multi}
Lin, S.-Y., C.-M. Huang, and T.-Y. Wu, Multi-access edge computing-based vehicle-vehicle-RSU data offloading over the multi-RSU-overlapped environment. \emph{IEEE Open Journal of Intelligent Transportation Systems}, Vol.~3, 2022, pp. 7--32.

\bibitem[{Ali et~al.(2016)Ali, Rahman, Chong, and Samantha}]{ali2016efficient}
Ali, G. M.~N., M.~A. Rahman, P.~H.~J. Chong, and S.~K. Samantha, On efficient data dissemination using network coding in multi-rsu vehicular ad hoc networks. In \emph{2016 IEEE 83rd Vehicular Technology Conference (VTC Spring)}, IEEE, 2016, pp. 1--5.

\bibitem[{Liu and Lee(2010)}]{liu2010rsu}
Liu, K. and V.~C. Lee, RSU-based real-time data access in dynamic vehicular networks. In \emph{13th International IEEE Conference on Intelligent Transportation Systems}, IEEE, 2010, pp. 1051--1056.

\bibitem[{Gao et~al.(2020)Gao, Liu, Li, and Yang}]{gao2020v2vr}
Gao, H., C.~Liu, Y.~Li, and X.~Yang, V2VR: reliable hybrid-network-oriented V2V data transmission and routing considering RSUs and connectivity probability. \emph{IEEE Transactions on Intelligent Transportation Systems}, Vol.~22, No.~6, 2020, pp. 3533--3546.

\bibitem[{Wu et~al.(2023)Wu, Cheng, Parker, Ran, and Noyce}]{wu2023development}
Wu, K., Y.~Cheng, S.~T. Parker, B.~Ran, and D.~A. Noyce, Development of the Data Pipeline for a Connected Vehicle Corridor. In \emph{International Conference on Transportation and Development 2023}, 2023, pp. 218--230.

\bibitem[{Li et~al.(2024)Li, Wu, Cheng, Parker, and Noyce}]{10292930}
Li, P., K.~Wu, Y.~Cheng, S.~T. Parker, and D.~A. Noyce, How Does C-V2X Perform in Urban Environments? Results From Real-World Experiments on Urban Arterials. \emph{IEEE Transactions on Intelligent Vehicles}, Vol.~9, No.~1, 2024, pp. 2520--2530.

\bibitem[{Pycrate(2024)}]{timmurphy.org2}
Pycrate, \emph{Pycrate}. [Online], 2024, \url{https://github.com/pycrate-org/pycrate}.

\bibitem[{Dosovitskiy et~al.(2017)Dosovitskiy, Ros, Codevilla, Lopez, and Koltun}]{Dosovitskiy17}
Dosovitskiy, A., G.~Ros, F.~Codevilla, A.~Lopez, and V.~Koltun, {CARLA}: {An} Open Urban Driving Simulator. In \emph{Proceedings of the 1st Annual Conference on Robot Learning}, 2017, pp. 1--16.

\end{thebibliography}
\end{document}